\pdfoutput=1

\documentclass[11pt]{article}

\usepackage[preprint]{acl}

\usepackage{times}
\usepackage{latexsym}

\usepackage[T1]{fontenc}

\usepackage[utf8]{inputenc}

\usepackage{microtype}

\usepackage{inconsolata}

\usepackage{graphicx}

\usepackage{caption}
\usepackage{subcaption}
\usepackage{tcolorbox}
\usepackage{xcolor}
\usepackage{listings}
\usepackage{booktabs}
\usepackage{threeparttable}
\usepackage{adjustbox}
\usepackage{rotating}
\usepackage{amsmath}
\usepackage{amssymb}
\usepackage{mathtools}
\usepackage{amsthm}
\usepackage{enumitem}

\lstdefinelanguage{json}{
    basicstyle=\ttfamily\tiny,
    showstringspaces=false,
    breaklines=true,
    stringstyle=\color{black},
    morestring=[b]",
    morecomment=[l]{//},
    commentstyle=\color{green},
    morekeywords={true, false, null}
}

\newcommand{\framework}{\textsc{Self-Discover}}
\newcommand{\iframework}{i\textsc{Self-Discover}}
\newcommand{\FaR}{\textsc{FaR}~}
\newcommand{\model}{\mathcal{M}}

\title{Effects of structure on reasoning in instance-level \framework}

\author{Sachith Gunasekara \and Yasiru Ratnayake \\
        Surge Global \\ 215 R A De Mel Mawatha, Colombo, Sri Lanka \\
        \texttt{\{\href{mailto:sachith@surge.global}{sachith}, \href{mailto:yasiru@surge.global}{yasiru}\}@surge.global} \\}

\begin{document}
\maketitle
\begin{abstract}
	The drive for predictable LLM reasoning in their integration with compound systems has popularized structured outputs, yet concerns remain about performance trade-offs compared to unconstrained natural language. At the same time, training on unconstrained Chain of Thought (CoT) traces has brought about a new class of strong reasoning models that nevertheless present novel compute budget and faithfulness challenges. This paper introduces \iframework, an instance-level adaptation of the \framework~framework, and using it compares dynamically generated structured JSON reasoning with its unstructured counterpart. Our empirical evaluation across diverse benchmarks using state-of-the-art open-source models supports a consistent advantage for unstructured reasoning. Notably, on the complex MATH benchmark, unstructured plans achieved relative performance improvements of up to 18.90\% over structured approaches. Zero-shot unstructured \iframework~variants are also shown to outperform their five-shot structured counterparts, underscoring the significance of this gap, even when structured plans are dynamically generated to ensure reasoning precedes the final answer. We further demonstrate that the optimal granularity of plan generation (instance-level vs. task-level) is context-dependent. These findings invite re-evaluation of the reliance on structured formats for complex problem-solving and how compound systems should be organized.
\end{abstract}

\section{Introduction}\label{sec:intro}
The pursuit of reliable, predictable and controllable outputs from Large Language Models (LLMs) in complex, multi-step reasoning tasks, especially for their integration into compound systems \citep{compound-ai-blog} through tool use/function calling (e.g., \citealp{OpenAI_FunctionCalling,schickToolformerLanguageModels2023}) has spurred the adoption of structured output formats, notably JSON. Concurrently, especially since the advent of CoT \citep{weiChainofthoughtPromptingElicits2024}, in-context learning and more generally test-time compute techniques have advanced so-called `System 2' reasoning capabilities in LLMs, especially in large models that have been specifically trained or fine-tuned particularly for these (whether with supervised fine-tuning or reinforcement learning) \citep{DBLP:journals/corr/abs-2501-12948}. While CoT traces are used in open models like DeepSeek R1 \citep{DBLP:journals/corr/abs-2501-12948}, this has led to a bifurcation of models between reasoning models and non-reasoning models based on application context due to undesirable behaviour of reasoning models such as overthinking \citep{StopOverthinkingSurvey2025}, faithfulness concerns \citep{chenReasoningModelsDont} and proneness to hallucination \citep{Vectara_HallucinationLeaderboard_Repo,Vectara_DeepseekHallucination_2024}. Techniques that go beyond basic CoT address some of these concerns. In particular, \framework~\citep{zhouSELFDISCOVERLargeLanguage2024a} and Foresee and Reflect (\FaR) \citep{zhouHowFaRAre2023} are examples of such techniques that further utilise schemas in how they work, albeit with distinct approaches to schema representation: \FaR typically utilizes a fixed JSON schema, while \framework~dynamically generates task-specific JSON structures. This leveraging of structure, especially with the latter approach's dynamism, while offering flexibility, offers an opportunity to put to test broader community positions about the efficacy of structured outputs.

Concerns have been raised that rigid, predefined template-based formats can degrade LLM performance~\citep{tam-etal-2024-speak}, and that suboptimal JSON schema design or implementation can hinder the very reasoning processes they aim to support~\citep{Castillo_GeminiStructuredOutputs}. Conversely, best practices emphasize the importance of well-formed structured outputs where reasoning steps clearly precede the final answer, a principle that, if adhered to, can enhance clarity and utility~\citep{DotTXT_SayWhatYouMean}. The reasoning process inherent in \framework, where modules are selected, adapted, and then implemented, naturally aligns with these principles of structured thought.

Our work extends this line of inquiry by comparing these dynamically generated JSON structures directly against unstructured, free-form natural language reasoning. To facilitate this investigation, we introduce \iframework, an instance-level adaptation of the \framework~framework. The primary motivation for this shift is to explore whether instance-specific adaptation of reasoning plans can unlock further performance enhancements, especially for benchmarks with diverse problem types where a single, task-level plan might prove suboptimal. Crucially, \iframework~is designed to accommodate the generation of both structured (dynamic JSON) and unstructured (natural language) reasoning traces for each specific problem instance, allowing for a direct comparison of reasoning styles at the same instance-level granularity. Our investigation, leveraging \iframework~capacity for instance-level generation across both reasoning styles, reveals a decisive advantage for unstructured natural language plans when pitted against their dynamically generated structured JSON counterparts, which are also produced on a per-instance basis. Notably, on the challenging MATH benchmark, this instance-level unstructured approach yields relative performance improvements of up to 18.90\% compared to the instance-level structured JSON alternative. Complementing this, on benchmarks such as BBH and T4D, our work addresses core questions:

\begin{enumerate}
	\item To what extent do unstructured natural language reasoning plans, generated on a per-instance basis, differ in performance compared to dynamically generated structured plans within a \framework-based framework, especially considering its alignment with principles of well-formed structured outputs?
	\item How does the granularity of reasoning plan generation (instance-level via \iframework~vs. task-level via the original \framework) impact performance across various benchmarks and language models?
	\item What is the influence of providing few-shot unlabeled task examples during the instance-level plan generation process?
\end{enumerate}

\section{Related Work}\label{sec:rel_work}
\subsection{Structured Reasoning}
\label{sec:rel_work-sub:structured_reasoning}

Recent work on structured reasoning has explored both fixed and dynamic schema generation. For instance, \textit{Foresee and Reflect} (\FaR) \citep{zhouHowFaRAre2023} utilizes a fixed JSON schema with a predefined template to solve the T4D benchmark. In contrast, \framework~\citep{zhouSELFDISCOVERLargeLanguage2024a} introduces a more adaptive methodology, allowing the language model to self-compose a task-specific JSON structure with dynamically `discovered' keys. This distinction between a fixed and a dynamic schema is illustrated in Figure~\ref{fig:json_distinction}.

Our study builds upon this \framework~framework but deviates in granularity and flexibility. While their work focuses on discovering a single task-level plan that is applied to all task instances, our \iframework~method generates a customized reasoning plan for each individual instance. Furthermore, it is designed to support not only structured JSON-based reasoning, but also an unstructured natural language counterpart, enabling a direct comparison between the two styles at the same instance-level granularity.

\begin{figure}[htbp]
	\centering
	\begin{subfigure}[t]{\columnwidth}
		\begin{tcolorbox}[
				colframe=lightgray,
				colback=lightgray,
				arc=2mm,
				boxrule=0.5mm,
				width=\columnwidth,
				fontupper=\small
			]
			\begin{lstlisting}[language=json]
{
  "Character A's likely future actions":
  "Potential challenge 1":
  "Can I help with it now by providing information?":
  "Potential challenge 2":
  "Can I help with it now by providing information? ":
  "Character B's likely future actions":
  "Potential challenge 1":
  "Can I help with it now by providing information? ":
  "final reasoning considering all steps above":
  "final answer":
}
			\end{lstlisting}
		\end{tcolorbox}
		\caption{The fixed JSON schema in the \FaR framework with its specific set of keys conforming to a predefined template}
		\label{subfig:json_far}
		\vspace{1mm}
	\end{subfigure}
	\begin{subfigure}[b]{\columnwidth}
		\begin{tcolorbox}[
				colframe=lightgray,
				colback=lightgray,
				arc=2mm,
				boxrule=0.5mm,
				width=\columnwidth,
				fontupper=\small
			]
			\begin{lstlisting}[language=json]
{
  "Identify the chain of events in the story":
  "Identify the consequences of each event":
  "Identify the cause-and-effect realtionships between events":
  "Choose a final answer based on the reasoning":
}
			\end{lstlisting}
		\end{tcolorbox}
		\caption{The task specific JSON structure in the \framework~framework with dynamically discovered keys}
		\label{subfig:json_self_discover}
	\end{subfigure}
	\caption{The different JSON structures employed by \FaR and \framework~demonstrated in \ref{subfig:json_far} and \ref{subfig:json_self_discover} respectively}
	\label{fig:json_distinction}
\end{figure}

Given that both \FaR and \framework~were evaluated on the T4D dataset \citep{zhouHowFaRAre2023}, which has not been made public, we implement their documented algorithm to transform the publicly available ToMi dataset \citep{le-etal-2019-revisiting} into the T4D format for our comparative analysis. Details can be found in Appendix~\ref{appendix:t4d}.

\subsection{Unstructured Reasoning}
\label{sec:rel_work-sub:unstructured_reasoning}

While foundational reasoning methods like Chain-of-Thought \citep{weiChainofthoughtPromptingElicits2024}, Self-Consistency \citep{DBLP:conf/iclr/0002WSLCNCZ23}, and Tree of Thoughts \citep{yaoTreeThoughtsDeliberate2023a}, among others \citep{zhouLargeLanguageModels2022,diao-etal-2024-active}, operate in an unstructured, free-form manner, there is a growing trend towards enforcing structured outputs for greater reliability, notably through features like JSON mode \citep{jsonModeOpenAI2024}. However, recent work has highlighted a potential performance trade-off. A study by \citet{tam-etal-2024-speak} claimed significant performance degradation when constraining LLM outputs, noting that natural language-based reasoning outperformed structured methods.

Crucially, the structured approach in their study used a simple, fixed template with keys like \textit{``step\_by\_step\_reasoning''}. This raises a key question: does this performance gap between structured and unstructured reasoning persist when the structured format is more sophisticated, such as the dynamically generated, task-specific schemas used in \framework? Our work directly addresses this question. By comparing free-form natural language reasoning against dynamically generated JSON plans within the same instance-level framework, we provide a more controlled investigation into the effects of structure on complex reasoning.

\section{Instance-Level Self-Discovery of Reasoning Plans for Problem Solving}\label{sec:phased_self-discover}
\begin{figure*}[ht]
	\centering
	\begin{subfigure}[t]{0.9\textwidth}
		\includegraphics[width=\textwidth]{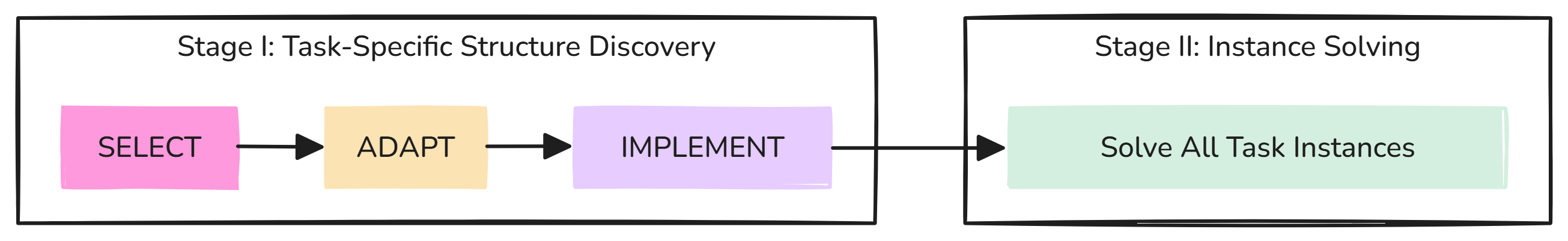}
		\caption{The original \framework~framework \citep{zhouSELFDISCOVERLargeLanguage2024a}: a two-stage process of task-level structure discovery and instance solving.}
		\label{subfig:self_discover}
		\vspace{1mm}
	\end{subfigure}
	\vspace{1mm} 
	\begin{subfigure}[b]{0.9\textwidth}
		\includegraphics[width=\textwidth]{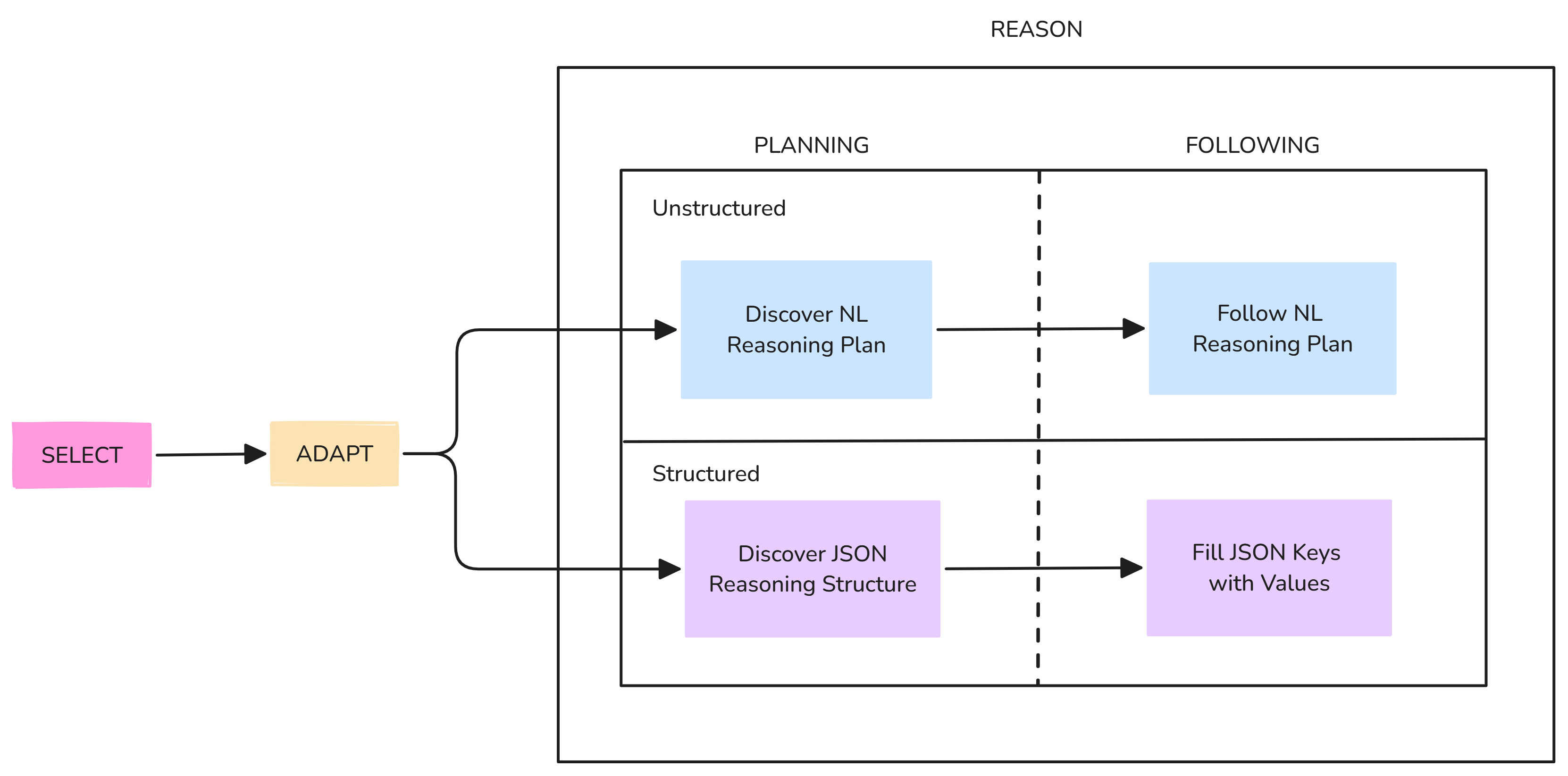}
		\caption{Our proposed \iframework: an instance-level workflow combining SELECT, ADAPT, and REASON (PLANNING + FOLLOWING) steps for each problem.}
		\label{subfig:phased_self_discover}
	\end{subfigure}
	\caption{Architectural comparison of (a) the original two-stage, task-level \framework~ and (b) our proposed instance-level \iframework. \iframework~generates and executes a custom plan for each instance, unlike \framework's batched discovery and application process.}
	\label{fig:framework_difference}
\end{figure*}

Our approach, \iframework, modifies the \framework~framework by generating a reasoning plan for each individual task instance. As illustrated in Figure~\ref{fig:framework_difference}, this can be executed via structured (JSON) or unstructured (natural language) reasoning paths. We formalize this process below, following a similar notation as in \citet{zhouSELFDISCOVERLargeLanguage2024a}.

Given a task instance $t_i$, a set of reasoning module descriptions $D$, and potentially $k$ unlabeled few-shot examples $E_{fs}^{(k)}$, \iframework~operates through three steps: SELECT, ADAPT, and REASON. Note that if $k=0$, no few-shot examples are used.

\paragraph{SELECT} The language model $\model$ identifies a relevant subset of reasoning modules $D_S$ from the full set $D$ for the given instance $t_i$. This is guided by a selection prompt $p_S$:
\begin{equation}
	D_S = \model(p_S \parallel D \parallel t_i \parallel E_{fs}^{(k)})
\end{equation}

\paragraph{ADAPT} The selected modules $D_S$ are then tailored to be more specific to $t_i$. An adaptation prompt $p_A$ guides the transformation into task-specific descriptions $D_A$:
\begin{equation}
	D_A = \model(p_A \parallel D_S \parallel t_i \parallel E_{fs}^{(k)})
\end{equation}

\paragraph{REASON} The adapted modules $D_A$ are used to generate and execute a reasoning plan for $t_i$. This consists of two sub-steps, PLANNING and FOLLOWING.
\subparagraph{1. PLANNING} From the adapted modules $D_A$, the model generates either an unstructured natural language plan $R_U$ or a structured JSON plan $R_S$, using dedicated planning prompts $p_{P_U}$ and $p_{P_S}$, respectively.
\begin{gather}
	R_U = \model(p_{P_U} \parallel D_A \parallel t_i \parallel E_{fs}^{(k)}) \\
	R_S = \model(p_{P_S} \parallel D_A \parallel t_i \parallel E_{fs}^{(k)})
\end{gather}

\subparagraph{2. FOLLOWING} With the reasoning plan established, the model executes it to derive a final answer. This step does not require the few-shot examples $E_{fs}^{(k)}$. Guided by prompts $p_{F_U}$ or $p_{F_S}$, the model follows the unstructured plan $R_U$ or the structured plan $R_S$ to produce the final answer, $A_U$ or $A_S$, respectively.
\begin{gather}
	A_U = \model(p_{F_U} \parallel R_U \parallel t_i) \\
	A_S = \model(p_{F_S} \parallel R_S \parallel t_i)
\end{gather}
The specific prompts ($p_S, p_A, p_{P_U}, p_{P_S}, p_{F_U}, p_{F_S}$) are detailed in Appendix~\ref{appendix:prompts}.

\section{Experiment Setup}\label{sec:experimental_design}
To rigorously evaluate our proposed \iframework, we conduct comprehensive experiments across multiple standard benchmarks. Our methodology focuses on comparing \iframework~with the original \framework~framework, and exploring the efficacy of its structured and unstructured reasoning capabilities under various guidance conditions.

\subsection{Evaluation Benchmarks}
\label{subsec:benchmarks}

\paragraph{BIG-Bench Hard (BBH)} \citep{suzgun-etal-2023-challenging}: A collection of 27 challenging tasks from BIG-Bench \citep{srivastava2023beyond}, encompassing algorithmic reasoning, natural language understanding, and world knowledge. Further details are provided in Appendix~\ref{appendix:bbh}.

\paragraph{Thinking for Doing (T4D)} \citep{zhouHowFaRAre2023}: A grounded social agent reasoning task requiring mental state inference. As the T4D dataset has not been publicly released, we replicated its creation methodology. Details of our replication process are available in Appendix~\ref{appendix:t4d}.

\paragraph{MATH} \citep{DBLP:conf/nips/HendrycksBKABTS21}: For this benchmark, we follow the precedent set by \framework~and subsample 200 examples from the official test set (see Appendix~\ref{appendix:math_sampling}). We highlight that the original \framework~also performs instance-level reasoning for this benchmark due to its complexity, hence rendering \framework~and \iframework~directly equivalent here.

\subsection{Models}
\label{subsec:models}
Our evaluations utilize prominent open-source LLMs: LLaMA-3.1-405B-Instruct (LLaMA) \cite{grattafioriLlama3Herd2024} and Mistral-Large (Mistral) \cite{AuLargeMistral}.

\subsection{Comparative Approaches}
\label{subsec:comparative_approaches}
Our experimental design is structured to first establish a strong baseline with \framework~and then to compare \iframework~under various configurations.

\subsubsection{Baseline}
Our primary baseline is \framework, selected for its demonstrated superiority over methods like Direct Prompting, Chain-of-Thought \cite{weiChainofthoughtPromptingElicits2024, kojimaLargeLanguageModels}, and Plan-and-Solve \cite{wangPlanandSolvePromptingImproving2023}. We apply its standard task-level approach to the BBH and T4D benchmarks, providing ten randomly selected unlabeled examples\footnote{The exact number used to discover a reasoning plan was not mentioned in the original paper; we selected 10 for our evaluations.} to discover a reasoning structure for each of the 27 BBH tasks and a single structure for the T4D benchmark.

A direct task-level comparison on the MATH benchmark is not appropriate, as the original \framework~paper itself employed instance-level reasoning. Consequently, our experiments on MATH focus exclusively on comparing structured versus unstructured reasoning within our proposed \iframework.

\subsubsection{\iframework}
We evaluate \iframework's efficacy through two sets of experiments designed to test its core capabilities and its response to contextual guidance.

\paragraph{A. Core Instance-Level Reasoning} First, we assess the fundamental performance of \iframework. We test both \textbf{\iframework~\textit{(Structured)}} and \textbf{\textit{(Unstructured)}} variants. For BBH and T4D, these are evaluated in a 0-shot setting ($E_{fs}^{(0)}$). For the MATH benchmark, following the same protocol in \framework, we guide plan generation with a single complete example from the training set, with further details on the selection and formatting provided in Appendix~\ref{appendix:MATH_formatting_one_shot}.

\paragraph{B. Few-Shot Contextual Guidance (BBH \& T4D only)} Next, to investigate whether instance-level planning can benefit from additional task context (analogous to the examples used in \framework's Stage 1), we introduce configurations guided by five unlabeled task examples ($E_{fs}^{(5)}$). This allows us to measure the impact of in-context learning on instance-specific plan generation. The two configurations are \textbf{\iframework~\textit{+ 5-Shot (Structured)}} and \textbf{\textit{+ 5-Shot (Unstructured)}}.

\subsection{Evaluation Metrics}
\label{subsec:evaluation_metrics}
We use accuracy as the evaluation metric across all benchmarks, maintaining consistency with prior work. For the BBH benchmark, we report the aggregate results across all 27 tasks (per-task results are in Appendix \ref{appendix:bbh_per_task_results}). For T4D and MATH, overall accuracy is reported. All improvements are calculated as Relative Change (RC) using the standard formula.

\section{Results}\label{sec:results}
Table~\ref{tab:main_results} presents the overall accuracy for all evaluated configurations. The subsequent analysis focuses on three key dimensions: the impact of reasoning style (structured vs. unstructured), the efficacy of plan generation granularity (instance vs. task-level), and the influence of few-shot guidance.

\begin{table*}[htbp!]
	\centering
	\caption{Overall performance (accuracy \%) of \iframework~variants compared to the \framework~(Baseline) across BBH (Average), T4D, and MATH benchmarks. BBH results are averaged over 27 tasks. For MATH, \framework~(Baseline) is instance-level structured, equivalent to our \iframework~(Struct, 0-shot). Few-shot guidance (5-shot) is not applied to MATH experiments.}
	\label{tab:main_results}
	\resizebox{\textwidth}{!}{%
		\begin{threeparttable}
			\begin{tabular}{@{}l|ccc|ccc@{}}
				\toprule
				                                    & \multicolumn{3}{c|}{\textbf{LLaMA}} & \multicolumn{3}{c}{\textbf{Mistral}}                                                                                 \\
				\textbf{Method}                     & \textbf{BBH}                        & \textbf{T4D}                         & \textbf{MATH (1-shot)} & \textbf{BBH} & \textbf{T4D} & \textbf{MATH (1-shot)} \\
				\midrule
				\framework~(Baseline)               & 86.59                               & 100.00                               & 63.50\tnote{\dag}      & 82.04        & 96.63        & 67.50\tnote{\dag}      \\
				\midrule
				\iframework~(Struct, 0-shot)   & 85.05                               & 73.23                                & 63.50                  & 83.14        & 76.06        & 67.50                  \\
				\iframework~(Unstruct, 0-shot) & 87.27                               & 78.90                                & 75.50                  & 85.57        & 82.09        & 76.50                  \\
				\midrule
				\iframework~(Struct, 5-shot)   & 85.14                               & 71.10                                & --                     & 84.22        & 78.90        & --                     \\
				\iframework~(Unstruct, 5-shot) & 87.02                               & 86.35                                & --                     & 86.29        & 88.29        & --                     \\
				\bottomrule
			\end{tabular}%
			\begin{tablenotes}
				\small 
				\item[\dag] This value is the same as the Structured zero-shot \iframework, and is duplicated here for clarity. No separate experiment was run for the MATH baseline.
			\end{tablenotes}
		\end{threeparttable}
	}
\end{table*}

\subsection{Impact of Reasoning Style: Structured vs. Unstructured}
\label{subsec:reasoning_style_impact}
Our findings consistently reveal a distinct advantage for the unstructured variant of \iframework~over its structured counterpart across various settings. This performance benefit is often so substantial that the 0-shot unstructured variant frequently surpasses its 5-shot guided structured counterpart. For instance, with LLaMA on T4D, the 0-shot unstructured approach outperformed the 5-shot structured alternative by a significant margin (10.97\% in relative terms). A similar pattern can be observed with BBH on the same model as well as Mistral on both BBH and T4D.

The MATH benchmark most dramatically underscores the potency of unstructured reasoning. Transitioning from the structured implementation to its unstructured counterpart boosted accuracy by a substantial 18.90\% relatively for LLaMA, while with Mistral, the unstructured approach delivered a notable relative gain of 13.33\%.

\subsection{Efficacy of Instance-Level vs. Task-Level Reasoning}
\label{subsec:instance_vs_task_level}
The results indicate that the optimal plan generation granularity (instance-level vs. task-level) is benchmark and model dependent. On the diverse BBH benchmark, instance-level reasoning showed clear benefits with Mistral, where the unstructured variant outperformed the task-level baseline by 4.30\%. With LLaMA, only the unstructured variant achieved a minor improvement of 0.79\%.

Conversely, on the coherent T4D benchmark, task-level \framework~exhibited markedly stronger performance. With LLaMA, the \framework~baseline (100.00\% accuracy) outperformed the 0-shot structured and unstructured \iframework~variants by 36.56\% and 26.74\%, respectively. A similar trend was observed with Mistral. These findings demonstrate no universal superiority between instance- and task-level reasoning; the most effective strategy appears contingent on task characteristics and the language model employed.

\subsection{Influence of Few-Shot Guidance on Instance-Level Plan Generation}
\label{subsec:few_shot_guidance_impact}
The influence of 5-shot guidance appears strongly correlated with the internal consistency of the benchmark's tasks. On the T4D benchmark, where task instances are highly similar, providing 5-shot guidance yielded consistent benefits. For unstructured plans, it led to notable relative improvements of 9.44\% with LLaMA and 7.55\% with Mistral.

In contrast, on the highly diverse BBH benchmark, the effect was marginal or even negative. With LLaMA, 5-shot guidance resulted in a slight performance decrease for unstructured plans, while Mistral showed only a modest increase. These results suggest that the utility of providing additional contextual examples for instance-level planning depends heavily on the diversity of problem types within a benchmark.

\section{Conclusion}\label{sec:conclusion}
The most striking finding of our study is the consistent performance advantage of unstructured natural language reasoning plans over dynamically generated structured JSON plans within \iframework. This superiority was particularly pronounced on the MATH benchmark and was further evidenced by 0-shot unstructured variants frequently outperforming their 5-shot structured counterparts on other benchmarks. These observations suggest a strong alignment of free-form text with models' pre-training, warranting further investigation into this performance gap.

Beyond this primary finding, our study revealed that the optimal reasoning strategy is highly context-dependent. Task-level reasoning demonstrated strength on a coherent benchmark like T4D, while instance-level plan generation often showed advantages on the diverse BBH. Similarly, the utility of few-shot guidance was not universal. The fact that it was most beneficial for coherent tasks like T4D, while its utility diminished for diverse tasks where generic examples might introduce noise, implies that for a truly instance-adaptive approach, a zero-shot configuration may be sufficient or even preferable. This suggests that a single instance can provide all necessary context without the risk of conflicting information from other examples.

In summary, this work demonstrates the significant performance benefits achievable by leveraging unstructured reasoning, even in instance-level self-discovery processes. It also highlights that choices regarding plan granularity and few-shot guidance are not universal but depend on task characteristics. In this current era of reasoning models being considered distinct from traditional LLMs rather than reasoning treated as a capability of a model, these findings around flexible, modular schemes for LLM thinking that can also be instance adaptive offer avenues for continued advancement of reasoning capabilities in large language models. They also make concrete the performance trade-off for structure vs free-form expression followed by structuring for practitioners building agentic compound systems around LLM capabilities, especially as industry converges on new protocols like MCP \citep{Anthropic_MCP_2024} and A2A \citep{Google_A2A_2024} for communication between agents and with tools.

\section{Limitations}\label{sec:limitations}
The comparison between structured and unstructured reasoning was primarily conducted within our \iframework~variants. The baseline was intentionally kept in its original form for a consistent comparison point, meaning that an unstructured variant of the task-level \framework~was not explored.

Secondly, our experiments utilized prominent open-source models. While these are powerful models, further benchmarking on state-of-the-art proprietary models is necessary to ascertain the generalizability of these findings.

Thirdly, the insights gathered about natural language reasoning against JSON are currently situated within the \iframework~family of methods. Their applicability and the extent of observed effects when applied to other reasoning frameworks or prompting strategies warrant further investigation.

\section{Future Work}\label{sec:future_work}
Building upon the findings and limitations of this study, several avenues for future research emerge. A primary direction is to delve deeper into the underlying reasons for the observed superiority of unstructured reasoning. This could involve analyses of model activations, studies on the cognitive load imposed by different output formats, or more controlled experiments on how LLMs translate thought processes into structured versus unstructured text.

Developing adaptive systems that can dynamically select the optimal plan granularity (task-level vs. instance-level) and reasoning style (structured vs. unstructured) based on the characteristics of the input query or task presents a promising avenue for robust LLM applications.

While unstructured reasoning showed strong performance, efforts to enhance structured reasoning are still crucial, especially for applications demanding high predictability, verifiability, and integration with downstream systems. This could involve designing more LLM-friendly or flexible JSON schemas, or developing improved training and fine-tuning strategies for structured output generation.

The practical impact of \iframework, particularly its unstructured variant, should be assessed by applying and evaluating it in complex, real-world agentic applications where multi-step reasoning is critical. This would also help understand its robustness and efficiency in more dynamic environments.


\bibliography{references}

\appendix


\section{Reproducing the T4D dataset}\label{appendix:t4d}

Given that the Thinking for Doing (T4D) dataset was not made publicly available, to conduct evaluations on the  benchmark, we replicate the methodology outlined in \cite{zhouHowFaRAre2023}. Leveraging the same ToMi dataset \cite{le-etal-2019-revisiting}, which focuses on Theory-of-Mind (ToM) inference questions, we filtered the ToMi examples to include only those involving ToM reasoning.

The implementation methodology involves:
\begin{enumerate}
	\item Identifying characters involved in the story scenario through templated narrative cue words: ``entered'', ``moved'', and ``exited''.
	\item Tracking the object of interest and character who moves the object by the cue word ``moved''.
\end{enumerate}

We make our replication publicly available as:

\begin{enumerate}
	\item \includegraphics[height=1em]{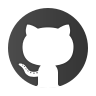} Code: The python script used to convert the ToMi dataset into T4D. (\url{https://github.com/sachith-gunasekara/t4d})
	\item \includegraphics[height=1em]{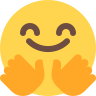} Dataset: The converted dataset using the above script. (\url{https://huggingface.co/datasets/sachithgunasekara/t4d})
\end{enumerate}



\section{Prompts for Reasoning Operations}\label{appendix:prompts}

This appendix details the high-level prompt structures for the core reasoning operations in \iframework. Figure~\ref{fig:common_prompts} shows the common architecture for the initial \textbf{SELECT} ($p_S$) and \textbf{ADAPT} ($p_A$) stages. Figure~\ref{fig:reason_prompts} details the distinct prompts for the \textbf{REASON} stage, which is composed of the \textbf{PLANNING} sub-step (using $p_{P_U}$ for unstructured and $p_{P_S}$ for structured plans) and the subsequent \textbf{FOLLOWING} sub-step (using $p_{F_U}$ and $p_{F_S}$, respectively).

\begin{figure*}[h!]
	\centering
	\includegraphics[width=\textwidth]{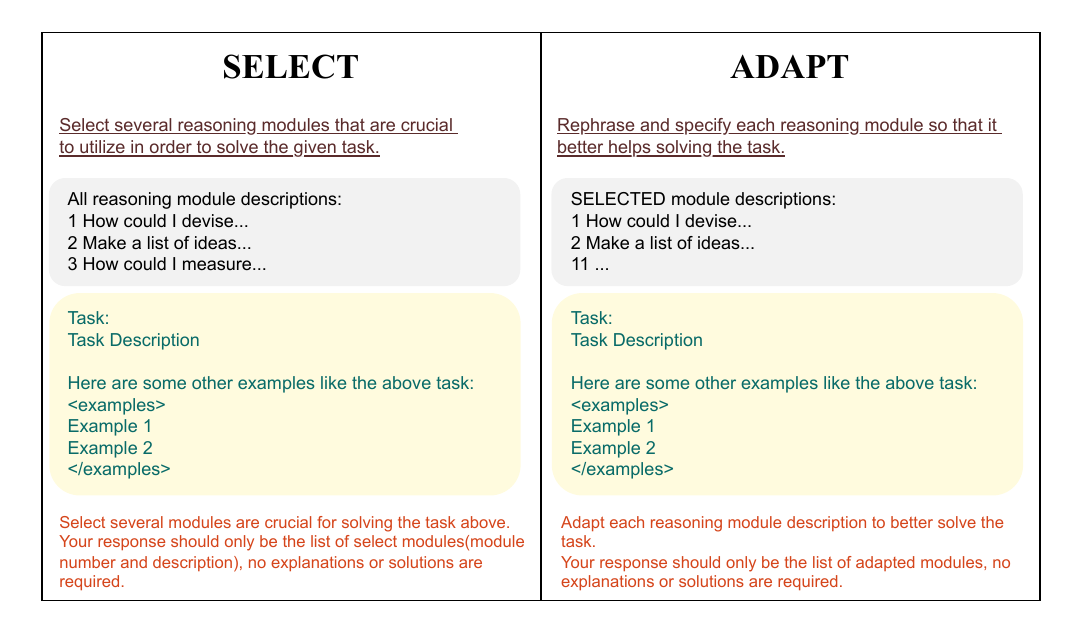}
	\caption{General prompt structure for the SELECT and ADAPT stages in \iframework. These prompts incorporate the task description and, optionally, few-shot examples.}
	\label{fig:common_prompts}
\end{figure*}

\begin{figure*}[h!]
	\centering
	\includegraphics[width=\textwidth]{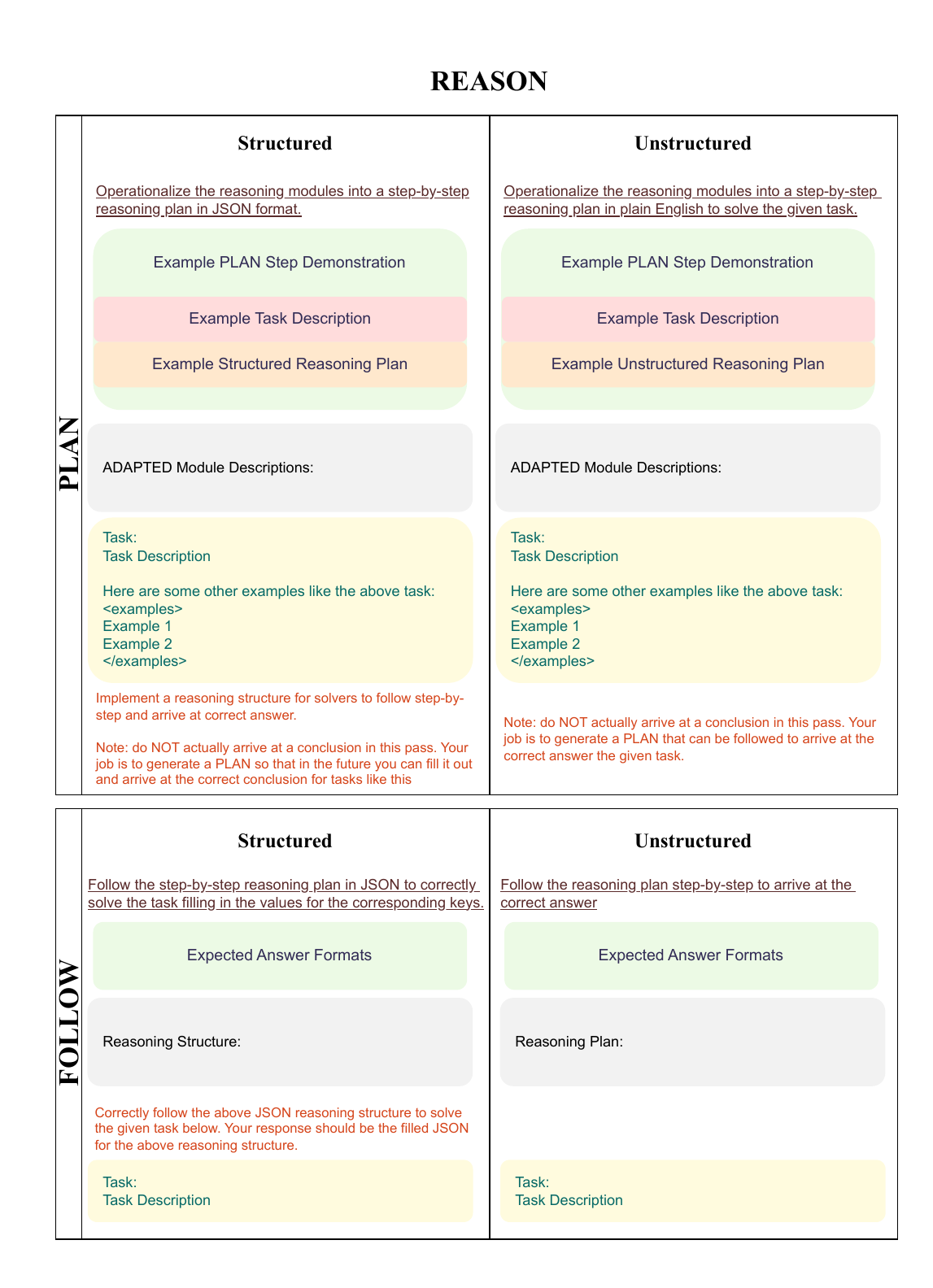}
	\caption{Prompt structure for the REASON stage (PLANNING and FOLLOWING sub-steps) in \iframework. Separate structures are shown for generating/following structured JSON plans and unstructured natural language plans. Placeholders for adapted modules, task descriptions, generated plans, and answer format instructions are filled contextually.}
	\label{fig:reason_prompts}
\end{figure*}

Key dynamic components within these prompt structures include:
\begin{itemize}
	\item \textbf{Task Description}: The specific problem instance, formatted according to the benchmark (BBH, T4D, MATH - with MATH including a one-shot example as detailed in Appendix~\ref{appendix:MATH_formatting_one_shot}).
	\item \textbf{Few-Shot Examples ($E_{fs}^{(k)}$)}: When $k>0$, unlabeled task examples are appended to the SELECT, ADAPT, and PLANNING prompts to provide additional context. These are omitted for 0-shot configurations and during the FOLLOWING sub-step.
	\item \textbf{Selected/Adapted Modules}: Outputs from the SELECT stage feed into ADAPT, and outputs from ADAPT feed into PLANNING.
	\item \textbf{Reasoning Plan}: The plan generated in the PLANNING sub-step is provided as input to the FOLLOWING sub-step.
	\item \textbf{Answer Format Instructions}: The FOLLOWING prompts include benchmark-specific guidelines on how the final answer should be presented.
\end{itemize}


\section{BBH Benchmark}\label{appendix:bbh}

The BIG-Bench Hard (BBH) benchmark is a collection of tasks derived from the Beyond the Imitation Game Benchmark (BIG-Bench). These tasks were specifically chosen because they proved particularly challenging for language models at the time of their selection, often requiring multi-step reasoning where model performance was near random. BBH aims to probe the limits of LLM capabilities in areas requiring deeper understanding and complex inference.

While the original BBH paper introduced 23 challenging tasks, the version of the dataset utilized in our experiments, sourced from \texttt{maveriq/bigbenchhard} on Hugging Face (\url{https://huggingface.co/datasets/maveriq/bigbenchhard}), includes a total of 27 distinct tasks. This expanded set features variations of tasks like logical deduction and object tracking with different object counts, such as \texttt{logical\_deduction\_three\_objects}, \texttt{logical\_deduction\_five\_objects}, \texttt{tracking\_shuffled\_objects\_three\_objects}, and \texttt{tracking\_shuffled\_objects\_five\_objects}.

Below is a list of the 27 BBH tasks included in the \texttt{maveriq/bigbenchhard} dataset used for our evaluations, along with a brief description for each:

\begin{enumerate}
	\item \textbf{\texttt{boolean\_expressions}}: Evaluates the model's ability to evaluate complex Boolean expressions.
	\item \textbf{\texttt{causal\_judgement}}: Assesses understanding of cause-and-effect relationships from textual descriptions.
	\item \textbf{\texttt{date\_understanding}}: Tests comprehension of dates, including relative dates and date arithmetic.
	\item \textbf{\texttt{disambiguation\_qa}}: Requires resolving ambiguities in questions to provide correct answers.
	\item \textbf{\texttt{dyck\_languages}}: Involves checking the validity of strings based on Dyck language rules (e.g., balanced parentheses).
	\item \textbf{\texttt{formal\_fallacies}}: Tests the ability to identify formal logical fallacies in arguments.
	\item \textbf{\texttt{geometric\_shapes}}: Assesses understanding of properties and relationships of geometric shapes.
	\item \textbf{\texttt{hyperbaton}}: Requires understanding sentences with inverted or non-standard word order (hyperbaton).
	\item \textbf{\texttt{logical\_deduction\_five\_objects}}: Tests deductive reasoning with statements involving five objects.
	\item \textbf{\texttt{logical\_deduction\_seven\_objects}}: Tests deductive reasoning with statements involving seven objects.
	\item \textbf{\texttt{logical\_deduction\_three\_objects}}: Tests deductive reasoning with statements involving three objects.
	\item \textbf{\texttt{movie\_recommendation}}: Assesses the ability to make movie recommendations based on preferences or descriptions.
	\item \textbf{\texttt{multistep\_arithmetic\_two}}: Involves solving multi-step arithmetic problems, often with two-digit numbers or two operations.
	\item \textbf{\texttt{navigate}}: Requires understanding and following navigational instructions (e.g., "If you take 2 steps forward, then 1 step left, are you at your starting point?").
	\item \textbf{\texttt{object\_counting}}: Tests the ability to count objects described in a text.
	\item \textbf{\texttt{penguins\_in\_a\_table}}: Involves reasoning about data presented in a tabular format, specifically about penguins.
	\item \textbf{\texttt{reasoning\_about\_colored\_objects}}: Tests reasoning about properties and relationships of colored objects.
	\item \textbf{\texttt{ruin\_names}}: Requires identifying "ruined" or slightly altered names.
	\item \textbf{\texttt{salient\_translation\_error\_detection}}: Tests the ability to detect salient errors in machine-translated text.
	\item \textbf{\texttt{snarks}}: Involves understanding and responding to "snarks" – sarcastic or subtly critical remarks, often posed as math or logic problems.
	\item \textbf{\texttt{sports\_understanding}}: Assesses comprehension of sports-related events, rules, and scenarios.
	\item \textbf{\texttt{temporal\_sequences}}: Requires understanding and reasoning about the order of events in time.
	\item \textbf{\texttt{tracking\_shuffled\_objects\_five\_objects}}: Tests the ability to track the positions of five objects that are shuffled.
	\item \textbf{\texttt{tracking\_shuffled\_objects\_seven\_objects}}: Tests the ability to track the positions of seven objects that are shuffled.
	\item \textbf{\texttt{tracking\_shuffled\_objects\_three\_objects}}: Tests the ability to track the positions of three objects that are shuffled.
	\item \textbf{\texttt{web\_of\_lies}}: Tests logical deduction to determine the truthfulness of individuals based on a set of interconnected statements where each person either tells the truth or lies.
	\item \textbf{\texttt{word\_sorting}}: Requires sorting a given list of words into alphabetical order.
\end{enumerate}


\section{Subsampling 200 Examples from the MATH Dataset}\label{appendix:math_sampling}

This section details the methodology used to subsample 200 examples from the MATH test dataset for our experiments, along with the statistics of this subsample.

\subsection{Source Dataset}
The source for our subsampling was the test split of the MATH dataset, originally comprising 5000 examples. We utilized a version equivalent to the \texttt{hendrycks/competition\_math} dataset, which was also the source for our training set examples (as detailed in Appendix~\ref{appendix:MATH_formatting_one_shot}). The original test set encompasses seven problem types: \textit{Algebra, Counting \& Probability, Geometry, Intermediate Algebra, Number Theory, Prealgebra, and Precalculus}.

\subsection{Subsampling Methodology}
To create a representative and manageable test set for our evaluations, we subsampled 200 examples from the 5000 available test instances. The primary goal of our subsampling strategy was to ensure that the distribution of problem types and difficulty levels within our 200-example subsample closely mirrored the proportions present in the original, larger test set.

The process involved:
\begin{enumerate}
	\item Analyzing the distribution of examples across different problem types (e.g., Algebra, Geometry) and difficulty levels (Level 1 to Level 5) in the full 5000-example test set.
	\item Performing a stratified sampling to select 200 examples such that the proportions of each problem type and each difficulty level were maintained as closely as possible to the original distributions.
\end{enumerate}

While we followed the precedent set by the \framework framework \cite{zhouSELFDISCOVERLargeLanguage2024a} in using a subsample of 200 examples, the original work did not specify details about maintaining proportional representation. Our stratified approach was adopted to enhance the representativeness of the subsample.

The specific 200-example subsample used in our experiments is publicly available at: \url{https://huggingface.co/datasets/sachithgunasekara/framework-MATH-subsample}.

\subsection{Statistics of the 200-Example Subsample}
The resulting subsample of 200 MATH problems has the following distributions by problem type and difficulty level:

\subsubsection{Distribution by Difficulty Level}
Table~\ref{tab:math_subsample_level} shows the number of problems from each difficulty level in our 200-example subsample.

\begin{table}[h!]
	\centering
	\small 
	\caption{Distribution of MATH Subsample by Difficulty Level}
	\label{tab:math_subsample_level}
	\begin{tabular}{lr}
		\hline
		\textbf{Difficulty Level} & \textbf{Number of Examples} \\
		\hline
		Level 1                   & 17                          \\
		Level 2                   & 36                          \\
		Level 3                   & 46                          \\
		Level 4                   & 49                          \\
		Level 5                   & 52                          \\
		\hline
		\textbf{Total}            & \textbf{200}                \\
		\hline
	\end{tabular}
\end{table}

\subsubsection{Distribution by Problem Type}
Table~\ref{tab:math_subsample_type} shows the number of problems from each problem type in our 200-example subsample.

\begin{table}[h!]
	\centering
	\small 
	\caption{Distribution of MATH Subsample by Problem Type}
	\label{tab:math_subsample_type}
	\begin{tabular}{lr}
		\hline
		\textbf{Problem Type}   & \textbf{Number of Examples} \\
		\hline
		Algebra                 & 47                          \\
		Intermediate Algebra    & 36                          \\
		Prealgebra              & 35                          \\
		Precalculus             & 22                          \\
		Number Theory           & 22                          \\
		Counting \& Probability & 19                          \\
		Geometry                & 19                          \\
		\hline
		\textbf{Total}          & \textbf{200}                \\
		\hline
	\end{tabular}
\end{table}

\subsubsection{Cross-Tabulation of Problem Type and Difficulty Level}
Table~\ref{tab:math_subsample_crosstab} provides a detailed breakdown of the 200 subsampled examples by both problem type and difficulty level.

\begin{table*}[h!]
	\centering
	\small 
	\caption{Cross-Tabulation of MATH Subsample by Problem Type and Difficulty Level}
	\label{tab:math_subsample_crosstab}
	\begin{tabular}{lrrrrr}
		\hline
		\textbf{Problem Type}   & \textbf{Level 1} & \textbf{Level 2} & \textbf{Level 3} & \textbf{Level 4} & \textbf{Level 5} \\
		\hline
		Algebra                 & 5                & 8                & 11               & 11               & 12               \\
		Counting \& Probability & 2                & 4                & 4                & 4                & 5                \\
		Geometry                & 2                & 3                & 4                & 5                & 5                \\
		Intermediate Algebra    & 2                & 5                & 8                & 10               & 11               \\
		Number Theory           & 1                & 4                & 5                & 6                & 6                \\
		Prealgebra              & 3                & 7                & 9                & 8                & 8                \\
		Precalculus             & 2                & 5                & 5                & 5                & 5                \\
		\hline
	\end{tabular}
\end{table*}


\section{Formatting MATH examples with a one-shot demonstration}\label{appendix:MATH_formatting_one_shot}

For the MATH benchmark, both structured and unstructured \iframework~variants utilize a one-shot demonstration to guide the plan generation and problem-solving process. This section details the selection and formatting of these one-shot examples.

\subsection{Source and Selection of One-Shot Examples}
The one-shot examples are drawn from the official MATH training dataset, specifically the \texttt{qwedsacf/competition\_math} dataset available on Hugging Face.

For each test instance from the MATH benchmark, a unique one-shot example (comprising a problem and its corresponding solution) is dynamically selected from this training set. The selection process is as follows:
\begin{enumerate}
	\item The ``level'' (e.g., ``Level 1'', ``Level 2'', etc.) and ``type'' (e.g., ``Algebra'', ``Number Theory'', etc.) of the current test instance are identified.
	\item The training dataset is filtered to find all examples that match the identified level and type.
	\item From this filtered subset, one example is randomly selected to serve as the one-shot demonstration for the current test instance.
\end{enumerate}
This instance-specific, dynamic selection of a relevant one-shot example mirrors the methodology employed by the original \framework~framework for the MATH benchmark.

\subsection{Formatting the One-Shot Demonstration}
The selected one-shot example is integrated directly into the task description provided to the LLM. The raw problem text and the complete, unprocessed solution text (including all reasoning steps) from the chosen training example are used.

The formatting follows the template below:
\begin{verbatim}
Problem: {problem}

<<<BEGIN: An example problem and solution>>>
Problem: {one_shot_example_problem}
Solution: {one_shot_example_solution}
<<<END: An example problem and solution>>>
\end{verbatim}
In this template:
\begin{itemize}
	\item \texttt{\{problem\}}: The current MATH test problem to be solved.
	\item \texttt{\{one\_shot\_example\_problem\}}: The problem statement from the selected one-shot training example.
	\item \texttt{\{one\_shot\_example\_solution\}}: The full solution from the selected one-shot training example.
\end{itemize}

\subsection{Usage in Prompts}
The complete string generated by the formatting described above (i.e., the current test problem combined with the selected one-shot problem and solution) is used as the value for the \texttt{\{task\_description\}} placeholder. This combined task description is consistently passed to the LLM throughout the following stages of the \iframework~process for the MATH benchmark:
\begin{itemize}
	\item SELECT
	\item ADAPT
	\item PLANNING
\end{itemize}
This ensures that the context of a solved example of similar type and difficulty is available to the model at the necessary steps.


\section{BBH Benchmark Results}\label{appendix:bbh_per_task_results}

This section presents the detailed per-task accuracy (\%) on the 27 subsets of the BBH benchmark for all evaluated models and methods. The "Average" row indicates the mean accuracy across all 27 subsets, which was also used as the overall BBH average in Section \ref{sec:results}. The evaluations from LLaMA and Mistral are shown in Tables \ref{tab:bbh_results_llama} and \ref{tab:bbh_results_mistral} respectively.

\begin{sidewaystable*}
	\centering
	\small 
	\caption{Per-task accuracy (\%) on BBH benchmark subsets for LLaMA. ``Struct'' refers to structured \iframework, and ``Unstruct'' refers to unstructured \iframework. ``\framework'' refers to the baseline.}
	\label{tab:bbh_results_llama}
	\begin{tabular}{@{}lrrrrr@{}}
		\toprule
		\textbf{Subset}                             & \shortstack{Self-                                                                     \\ Discover} & \shortstack{\iframework \\ (Struct, 0-shot)} & \shortstack{\iframework \\ (Unstruct, 0-shot)} & \shortstack{\iframework \\ (Struct, 5-shot)} & \shortstack{\iframework \\ (Unstruct, 5-shot)} \\
		\midrule
		boolean\_expressions                        & 97.60             & 98.40          & 98.00          & 98.40          & 98.00          \\
		causal\_judgement                           & 71.12             & 65.78          & 71.12          & 68.98          & 67.91          \\
		date\_understanding                         & 94.00             & 87.60          & 91.20          & 90.00          & 88.00          \\
		disambiguation\_qa                          & 79.60             & 78.80          & 76.40          & 75.20          & 74.40          \\
		dyck\_languages                             & 43.60             & 49.20          & 50.00          & 54.40          & 51.20          \\
		formal\_fallacies                           & 70.80             & 73.60          & 80.80          & 75.20          & 84.00          \\
		geometric\_shapes                           & 62.40             & 58.80          & 68.80          & 53.60          & 60.00          \\
		hyperbaton                                  & 98.40             & 90.80          & 91.20          & 92.40          & 94.00          \\
		logical\_deduction\_five\_objects           & 86.80             & 91.60          & 91.60          & 90.40          & 92.40          \\
		logical\_deduction\_seven\_objects          & 82.40             & 81.20          & 86.40          & 81.20          & 82.80          \\
		logical\_deduction\_three\_objects          & 100.00            & 98.00          & 99.20          & 99.20          & 99.60          \\
		movie\_recommendation                       & 68.80             & 66.80          & 68.00          & 68.80          & 66.80          \\
		multistep\_arithmetic\_two                  & 91.60             & 91.20          & 93.60          & 87.20          & 93.60          \\
		navigate                                    & 94.00             & 98.40          & 98.80          & 96.40          & 96.80          \\
		object\_counting                            & 97.20             & 86.00          & 97.20          & 89.20          & 97.20          \\
		penguins\_in\_a\_table                      & 94.52             & 96.58          & 97.95          & 96.58          & 97.95          \\
		reasoning\_about\_colored\_objects          & 91.60             & 92.40          & 92.00          & 96.80          & 96.00          \\
		ruin\_names                                 & 88.40             & 87.60          & 88.00          & 86.40          & 88.40          \\
		salient\_translation\_error\_detection      & 72.40             & 71.20          & 69.60          & 65.60          & 73.20          \\
		snarks                                      & 87.08             & 84.83          & 87.64          & 90.45          & 87.64          \\
		sports\_understanding                       & 86.80             & 70.80          & 79.60          & 75.20          & 82.00          \\
		temporal\_sequences                         & 99.60             & 99.60          & 99.20          & 99.20          & 100.00         \\
		tracking\_shuffled\_objects\_five\_objects  & 99.60             & 98.00          & 99.20          & 96.80          & 99.60          \\
		tracking\_shuffled\_objects\_seven\_objects & 98.40             & 98.40          & 98.80          & 98.40          & 98.40          \\
		tracking\_shuffled\_objects\_three\_objects & 98.80             & 99.60          & 99.20          & 97.60          & 99.20          \\
		web\_of\_lies                               & 91.20             & 88.80          & 91.60          & 84.80          & 88.00          \\
		word\_sorting                               & 91.20             & 92.40          & 91.20          & 90.40          & 92.40          \\
		\midrule
		\textbf{Average}                            & \textbf{86.59}    & \textbf{85.05} & \textbf{87.27} & \textbf{85.14} & \textbf{87.02} \\
		\bottomrule
	\end{tabular}
\end{sidewaystable*}

\begin{sidewaystable*}
	\centering
	\small 
	\caption{Per-task accuracy (\%) on BBH benchmark subsets for Mistral. ``Struct'' refers to structured \iframework, and ``Unstruct'' refers to unstructured \iframework. ``\framework'' refers to the baseline.}
	\label{tab:bbh_results_mistral}
	\begin{tabular}{@{}lrrrrr@{}}
		\toprule
		\textbf{Subset}                             & \shortstack{Self-                                                                     \\ Discover} & \shortstack{\iframework \\ (Struct, 0-shot)} & \shortstack{\iframework \\ (Unstruct, 0-shot)} & \shortstack{\iframework \\ (Struct, 5-shot)} & \shortstack{\iframework \\ (Unstruct, 5-shot)} \\
		\midrule
		boolean\_expressions                        & 93.60             & 95.60          & 97.20          & 96.40          & 97.60          \\
		causal\_judgement                           & 73.26             & 70.05          & 71.66          & 73.26          & 71.12          \\
		date\_understanding                         & 84.40             & 81.60          & 80.00          & 81.20          & 84.40          \\
		disambiguation\_qa                          & 63.20             & 69.20          & 72.80          & 71.60          & 68.40          \\
		dyck\_languages                             & 58.00             & 42.80          & 50.00          & 50.40          & 53.20          \\
		formal\_fallacies                           & 75.60             & 72.40          & 82.40          & 72.80          & 79.20          \\
		geometric\_shapes                           & 47.20             & 66.40          & 71.20          & 58.80          & 68.00          \\
		hyperbaton                                  & 94.40             & 94.40          & 98.00          & 94.40          & 98.00          \\
		logical\_deduction\_five\_objects           & 82.80             & 88.80          & 94.00          & 92.40          & 92.80          \\
		logical\_deduction\_seven\_objects          & 81.20             & 78.80          & 89.60          & 81.60          & 87.60          \\
		logical\_deduction\_three\_objects          & 89.60             & 98.40          & 99.60          & 98.80          & 100.00         \\
		movie\_recommendation                       & 68.00             & 71.60          & 68.80          & 71.60          & 69.60          \\
		multistep\_arithmetic\_two                  & 89.20             & 84.80          & 92.00          & 86.80          & 89.60          \\
		navigate                                    & 91.60             & 93.60          & 96.80          & 92.40          & 96.80          \\
		object\_counting                            & 95.60             & 84.80          & 97.60          & 90.80          & 95.60          \\
		penguins\_in\_a\_table                      & 89.73             & 97.95          & 97.95          & 98.63          & 99.32          \\
		reasoning\_about\_colored\_objects          & 96.80             & 96.80          & 97.20          & 96.00          & 96.80          \\
		ruin\_names                                 & 74.40             & 76.00          & 74.00          & 74.80          & 76.40          \\
		salient\_translation\_error\_detection      & 66.40             & 65.60          & 68.00          & 65.60          & 70.00          \\
		snarks                                      & 88.20             & 85.96          & 87.08          & 87.08          & 89.33          \\
		sports\_understanding                       & 85.60             & 74.80          & 78.40          & 78.00          & 81.20          \\
		temporal\_sequences                         & 99.60             & 99.20          & 97.20          & 97.60          & 99.20          \\
		tracking\_shuffled\_objects\_five\_objects  & 75.60             & 97.60          & 98.40          & 96.80          & 100.00         \\
		tracking\_shuffled\_objects\_seven\_objects & 76.00             & 96.80          & 94.40          & 97.60          & 98.80          \\
		tracking\_shuffled\_objects\_three\_objects & 96.00             & 96.40          & 99.20          & 97.80          & 100.00         \\
		web\_of\_lies                               & 99.20             & 86.40          & 89.20          & 90.80          & 98.00          \\
		word\_sorting                               & 80.00             & 78.00          & 67.60          & 80.00          & 68.80          \\
		\midrule
		\textbf{Average}                            & \textbf{82.04}    & \textbf{83.14} & \textbf{85.57} & \textbf{84.22} & \textbf{86.29} \\
		\bottomrule
	\end{tabular}
\end{sidewaystable*}


\section{List of Base Reasoning Modules}\label{appendix:reasoning_modules}

The 39 reasoning modules provided to the LLM during the SELECT stage of the iSELF-DISCOVER and SELF-DISCOVER frameworks are given in Figure~\ref{box:reasoning_modules}. The model is tasked with selecting a subset of these modules that are most relevant for solving the given task or task examples.

\begin{figure*}[htp!]
	\centering
	\fbox{
		\begin{minipage}{\dimexpr0.95\linewidth-2\fboxsep-2\fboxrule\relax}
			\small
			\begin{enumerate}[nosep, label=\arabic*.]
				\item How could I devise an experiment to help solve that problem?
				\item Make a list of ideas for solving this problem, and apply them one by one to the problem to see if any progress can be made.
				\item How could I measure progress on this problem?
				\item How can I simplify the problem so that it is easier to solve?
				\item What are the key assumptions underlying this problem?
				\item What are the potential risks and drawbacks of each solution?
				\item What are the alternative perspectives or viewpoints on this problem?
				\item What are the long-term implications of this problem and its solutions?
				\item How can I break down this problem into smaller, more manageable parts?
				\item Critical Thinking: This style involves analyzing the problem from different perspectives, questioning assumptions, and evaluating the evidence or information available. It focuses on logical reasoning, evidence-based decision-making, and identifying potential biases or flaws in thinking.
				\item Try creative thinking, generate innovative and out-of-the-box ideas to solve the problem. Explore unconventional solutions, thinking beyond traditional boundaries, and encouraging imagination and originality.
				\item Seek input and collaboration from others to solve the problem. Emphasize teamwork, open communication, and leveraging the diverse perspectives and expertise of a group to come up with effective solutions.
				\item Use systems thinking: Consider the problem as part of a larger system and understanding the interconnectedness of various elements. Focuses on identifying the underlying causes, feedback loops, and interdependencies that influence the problem, and developing holistic solutions that address the system as a whole.
				\item Use Risk Analysis: Evaluate potential risks, uncertainties, and tradeoffs associated with different solutions or approaches to a problem. Emphasize assessing the potential consequences and likelihood of success or failure, and making informed decisions based on a balanced analysis of risks and benefits.
				\item Use Reflective Thinking: Step back from the problem, take the time for introspection and self-reflection. Examine personal biases, assumptions, and mental models that may influence problem-solving, and being open to learning from past experiences to improve future approaches.
				\item What is the core issue or problem that needs to be addressed?
				\item What are the underlying causes or factors contributing to the problem?
				\item Are there any potential solutions or strategies that have been tried before? If yes, what were the outcomes and lessons learned?
				\item What are the potential obstacles or challenges that might arise in solving this problem?
				\item Are there any relevant data or information that can provide insights into the problem? If yes, what data sources are available, and how can they be analyzed?
				\item Are there any stakeholders or individuals who are directly affected by the problem? What are their perspectives and needs?
				\item What resources (financial, human, technological, etc.) are needed to tackle the problem effectively?
				\item How can progress or success in solving the problem be measured or evaluated?
				\item What indicators or metrics can be used?
				\item Is the problem a technical or practical one that requires a specific expertise or skill set? Or is it more of a conceptual or theoretical problem?
				\item Does the problem involve a physical constraint, such as limited resources, infrastructure, or space?
				\item Is the problem related to human behavior, such as a social, cultural, or psychological issue?
				\item Does the problem involve decision-making or planning, where choices need to be made under uncertainty or with competing objectives?
				\item Is the problem an analytical one that requires data analysis, modeling, or optimization techniques?
				\item Is the problem a design challenge that requires creative solutions and innovation?
				\item Does the problem require addressing systemic or structural issues rather than just individual instances?
				\item Is the problem time-sensitive or urgent, requiring immediate attention and action?
				\item What kinds of solution typically are produced for this kind of problem specification?
				\item Given the problem specification and the current best solution, have a guess about other possible solutions.
				\item Let's imagine the current best solution is totally wrong, what other ways are there to think about the problem specification?
				\item What is the best way to modify this current best solution, given what you know about these kinds of problem specification?
				\item Ignoring the current best solution, create an entirely new solution to the problem.
				\item Let's think step by step.
				\item Let's make a step by step plan and implement it with good notion and explanation.
			\end{enumerate}
		\end{minipage}
	}
	\caption{List of the 39 Base Reasoning Modules.}
	\label{box:reasoning_modules} 
\end{figure*}

\end{document}